\documentclass[
]{ceurart}

\usepackage{longtable}
\usepackage{caption}
\usepackage{graphicx}
\DeclareCaptionFormat{myformat}{\normalsize\textbf{#1}\\\normalsize#3}
\begin{document}

\copyrightyear{2021}
\copyrightclause{Copyright for this paper by its authors.
  Use permitted under Creative Commons License Attribution 4.0
  International (CC BY 4.0).}

\conference{De-Factify 3.0: Third Workshop on Multimodal Fact Checking and Hate Speech Detection, co-located with AAAI 2024. Vancouver, Canada.}

\title{Team Trifecta at Factify5WQA: Setting the Standard in Fact Verification with Fine-Tuning}

\author{Shang-Hsuan Chiang†}[%
orcid=0009-0006-2001-8365,
email=andy10801@gmail.com,
url=https://andychiangsh.github.io/AndyChiangSH/,
]

\author{Ming-Chih Lo†}[%
orcid=0009-0005-8949-2770,
email=max230620089@gmail.com,
]

\author{Lin-Wei Chao†}[%
orcid=0009-0003-5460-8476,
email=william09172000@gmail.com,
]

\author{Wen-Chih Peng}[%
email=wcpeng@cs.nycu.edu.tw
]

\address{Department of Computer Science, National Yang Ming Chiao Tung University, Hsinchu, Taiwan}

\newcommand\blfootnote[1]{%
  \begingroup
  \renewcommand\thefootnote{}\footnote{#1}%
  \addtocounter{footnote}{-1}%
  \endgroup
}

\blfootnote{† These authors contributed equally to this work and share the first authorship.}

\begin{abstract}
In this paper, we present \textbf{Pre-CoFactv3}, a comprehensive framework comprised of Question Answering and Text Classification components for fact verification. Leveraging In-Context Learning, Fine-tuned Large Language Models (LLMs), and the FakeNet model, we address the challenges of fact verification. Our experiments explore diverse approaches, comparing different Pre-trained LLMs, introducing FakeNet, and implementing various ensemble methods. Notably, our team, Trifecta, secured \textbf{first place} in the \textbf{AAAI-24 Factify 3.0 Workshop} \footnote{https://defactify.com/factify3.html}, surpassing the baseline accuracy by 103\% and maintaining a 70\% lead over the second competitor. This success underscores the efficacy of our approach and its potential contributions to advancing fact verification research.
\end{abstract}


\maketitle

\section{Introduction}
In an era characterized by an overwhelming influx of information facilitated by the internet and social media, the verification of facts has emerged as an increasingly critical challenge. The proliferation of digital platforms has democratized content creation and dissemination, yet it has also paved the way for the rapid spread of misinformation, disinformation, and misleading content. This phenomenon poses a significant threat to the integrity of information consumed by individuals, communities, and societies at large.

The need for robust fact verification mechanisms has become increasingly evident. The ability to distinguish between accurate, credible information and falsehoods has become essential in ensuring informed decision-making, fostering a well-informed citizenry, and preserving the fabric of democratic societies. Fact verification serves as a cornerstone in the pursuit of truth, objectivity, and reliability in an information landscape fraught with distortions and inaccuracies.

In recent years, the advent of large language models has revolutionized natural language understanding, enabling machines to comprehend and generate human-like text at an unprecedented scale. Among the myriad applications of these models, one paramount challenge they face is the discernment of factual accuracy in claims against available evidence, due to the inherent limitations and complexities of language understanding. Thus we undertake the task of classifying claims as supported, refuted, or neutral based on provided evidence, advancing the capabilities of language models towards more nuanced and accurate comprehension.

This research paper explores the intricate process of enhancing large language models to navigate the terrain of claim verification, employing a spectrum of methodologies. Fine-tuning, recognized for its superior performance, serves as a cornerstone method in our investigation. This paper goes beyond by delving into the significance of in-context learning, which we consider to also encompass the wider spectrum of prompt engineering and prompt tuning, elucidating its role in augmenting models' understanding of nuanced linguistic contexts. Additionally, it examines the effectiveness and limitations of feature extraction techniques in capturing crucial information relevant to claim classification. Furthermore, the paper explores the benefits of ensemble learning approaches, synthesizing diverse model outputs to enhance classification accuracy and reliability.

Our approach draws substantial influence from the pioneering work of Du et al.'s Pre-CoFactv2 model \cite{du2023team}, as presented during the Factify 2 challenge \cite{suryavardan2023factify}. We were impressed by the performance of parameter-efficient fine-tuning on the DeBERTa model, and thought to take one step further, dedicating our efforts to a comprehensive fine-tuning of the advanced DeBERTaV3 \cite{he2021debertav3}. The outcome showcases substantial improvement across question answering and text classification tasks when compared to all other assessed methodologies. We hereby introduce our resultant model as \textbf{Pre-CoFactv3}, symbolizing our continuum of innovation derived from our predecessors’ pioneering endeavors.

Judging by evaluation results, our fine-tuning method outperforms in-context learning and human baseline, when discerning between support, refute and neutral correlations between claim and evidence text, reaching 86 percent accuracy overall on internal validation. When using external validation against other models, our method comes out as state-of-the-art, leading ahead of the next contender by 52 percent.

By synthesizing insights from varied methodologies and their contributions, this research aims to contribute significantly to the ongoing discourse on refining large language models for factual inference, paving the way for more robust and reliable natural language understanding in the domain of claim verification.

\section{Related Works}

\subsection{Traditional Methods}
In the landscape of fake news detection, traditional methods relied on rule-based approaches, focusing on specific words or phrases linked to misinformation. Models such as Naïve Bayes, Support Vector Machines (SVM), and Decision Trees exemplify these conventional techniques \cite{castillo2011information}\cite{zhao2015enquiring}. However, their vulnerability to phrasing variations and evolving tactics limited their effectiveness.

\subsection{Deep Learning Revolution in NLP}
The Deep Learning Revolution in Natural Language Processing (NLP) introduced advanced feature engineering and integrated sophisticated techniques. Notable methodologies like Convolutional Neural Networks (CNN), Recurrent Neural Networks (RNN), and Long Short-Term Memory (LSTM) marked this transformative phase. For instance, Kaliyar et al. \cite{kaliyar2020fndnet} harnessed deep CNN for multi-layered feature extraction, while Bahad et al. \cite{bahad2019fake} introduced the bi-directional LSTM-RNN architecture with the GLoVe word embedding for effective fake news detection.

\subsection{Cross-Modality}
To address the dynamic nature of misinformation, researchers explored cross-domain learning, integrating NLP knowledge from related domains. This led to the examination of multi-modal data analysis, incorporating text, images, and social media network information. Models like EANN (event adversarial neural network) by Wang et al. \cite{wang2018eann} and MCAN (multimodal co-attention network) by Wu et al. \cite{wu2021multimodal} exemplify these cross-modal approaches. However, due to the unimodal nature of the Factify5WQA dataset \cite{surya2024factify} (text only), our focus remains on unimodal models.

\subsection{The Power of Pre-trained Large Language Models}
In recent times, the evolution of Transformer \cite{vaswani2017attention} has revolutionized the landscape of natural language processing and information extraction. Pre-trained Large Language Models, exemplified by BERT \cite{devlin2018bert} and GPT \cite{brown2020language}, reveal profound capabilities in comprehending and generating human-like text, proving particularly adept at identifying fake news.
In line with the methodology outlined by Du et al. \cite{du2023team}, we leverage the simplicity and effectiveness inherent in various pre-trained models. Our architecture systematically tests different pre-trained models, thereby amplifying efficiency in training and learning patterns from the dataset.

\section{Methods}

\subsection{Pre-CoFactv3 Overview}

\begin{figure}
  \centering
  \includegraphics[width=\linewidth]{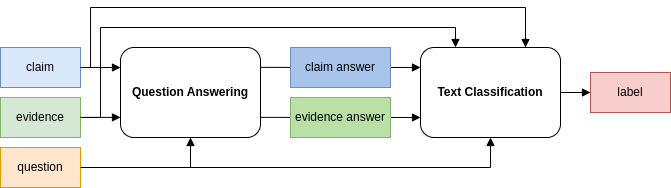}
  \caption{The overview of our Pre-CoFactv3 framework}
  \label{fig:overview}
\end{figure}


Figure \ref{fig:overview} illustrates the overview of our \textbf{Pre-CoFactv3} framework. Owing to the absence of claim answers, evidence answers, and labels in the testing dataset, we have divided the overall process into two distinct phases: (1) Question Answering and (2) Text Classification. In the first phase, questions are answered by the information derived from both the claim and evidence. Subsequently, in the second phase, the system uses the claim, evidence, and answers obtained from the initial phase to predict the appropriate label, which can be categorized as Support, Neutral, or Refute.

\subsection{Question Answering}


In the Question Answering phase, the input comprises a set denoted by $I = \{C_i, E_i, Q_i\}_{i=0}^{|I|}$, where $Q_i = \{Q_{ij}\}_{j=0}^{|Q|}$. Here, $C_i$ represents the claim, $E_i$ denotes the evidence, and $Q_i$ signifies a list of questions associated with the given claim and evidence. The output is represented by a set denoted as $O = \{CA_i, EA_i\}_{i=0}^{|O|}$, where $CA_i = \{CA_{ij}\}_{j=0}^{|CA|}$ and $EA_i = \{EA_{ij}\}_{j=0}^{|EA|}$. In this context, $CA_i$ corresponds to the list of claim answers, and $EA_i$ represents the list of evidence answers generated by the Question Answering module.


Within the Question Answering module, we employ two distinct methodologies: In-Context Learning and Fine-tuning Large Language Models (LLMs).

Our initial experimentation focuses on In-Context Learning, with detailed insights provided in Section \ref{In-Context Learning Question Answering}.

\subsubsection{Fine-tuning Large Language Models (LLMs)}

In the Fine-tuning of LLMs, we integrate two sets of LLMs. The first set involves LLMs fine-tuned using the SQuAD 2.0 dataset \cite{2016arXiv160605250R}, accessible on Hugging Face. Concurrently, the second set comprises pre-trained LLMs subsequently fine-tuned on the FACTIFY5WQA dataset \cite{surya2024factify}, specifically tailored for the question-answering task. The objective of the question-answering task is to input the context (claim or evidence) and subsequently identify the index corresponding to the location of the answer within that context. The formulation is as below:

\begin{equation}
\begin{aligned}
index\:of\:CA_{ij}\:in\:C_i = LLM(C_i, Q_{ij}) \\
index\:of\:EA_{ij}\:in\:E_i = LLM(E_i, Q_{ij})
\end{aligned}
\end{equation}

Comprehensive experimental results will be presented in Section \ref{Question Answering}.

\subsection{Text Classification}


In the Text Classification phase, the input is represented as $I = \{C_i, E_i, Q_i, CA_i, EA_i\}_{i=0}^{|I|}$, encompassing the claim $C_i$, evidence $E_i$, question $Q_i$, and the claim answer $CA_i$ and evidence answer $EA_i$ obtained from the preceding Question Answering phase. The output of the Text Classification module is the predicted label, which falls into one of the categories: Support, Neutral, or Refute.

Within the Text Classification module, three distinct methodologies are employed: In-Context Learning, FakeNet, and Fine-tuning Large Language Models (LLMs).

Our initial experimentation centers around In-Context Learning, and comprehensive insights into this approach are elucidated in Section \ref{In-Context Learning Text Classification}.

\subsubsection{FakeNet}

\begin{figure}
  \centering
  \includegraphics[width=\linewidth]{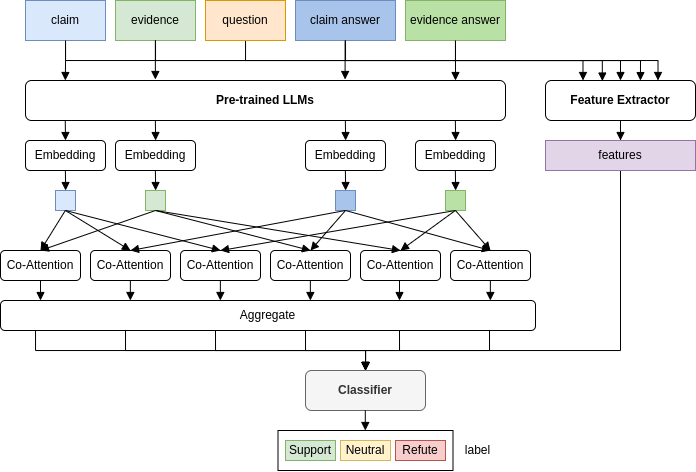}
  \caption{The overview of FakeNet}
  \label{fig:FakeNet}
\end{figure}

Building upon the foundations laid by the previous work, Pre-CoFactv2 \cite{du2023team}, we have introduced a novel framework named FakeNet. Figure \ref{fig:FakeNet} provides an overview of FakeNet.

\paragraph{Pre-trained LLMs}

We leverage Pre-trained Large Language Models (LLMs) to generate embeddings for the claim, evidence, claim answer, and evidence answer. Subsequently, we freeze the parameters of the Pre-trained LLMs and exclusively train the last embedding layer.

In the subsequent step, we employ four embeddings that participate in co-attention with each other in pairs, resulting in six pairs of co-attention. These co-attention pairs comprise (1) claim and evidence, (2) claim and claim answer, (3) claim and evidence answer, (4) evidence and claim answer, (5) evidence and evidence answer, and (6) claim answer and evidence answer. The co-attention block utilized here is a variant of the multi-head self-attention block, akin to the encoder in Transformer \cite{vaswani2017attention}. This block accepts two embeddings as inputs to facilitate the learning of interactions and relations between them. Subsequently, mean aggregation is employed to merge all the outcomes into a single embedding $E_{PLM}$, representing the corresponding text.

\paragraph{Feature Extractor}


Informed by the insights of Gao et al. \cite{gao2021logically}, our model leverages the length of the text and other common textual attributes in NLP to extract important information from diverse perspectives. Additionally, inspired by Zhang et al. \cite{zhang2023ino}, we incorporated the calculation of similarity between text and question-answer pairs, aiming to harness the significance of relevance and coherence to enhance our model's capabilities. The details of our model's feature extractors are illustrated in Table \ref{tab:Feature Selection}.

\begin{table}[]
\begin{tabular}{@{}clll@{}}
\toprule
\multicolumn{1}{c}{\textbf{Type}} & \multicolumn{3}{c}{\textbf{Features}} \\ \midrule
\multicolumn{1}{c|}{\multirow{4}{*}{\begin{tabular}[c]{@{}c@{}}Common Features \\ in NLP\end{tabular}}}   & Character count        & Word count            & Count of capital characters \\
\multicolumn{1}{c|}{}                                                                                     & Count of capital words & Count of punctuation & Count of words in quotes    \\
\multicolumn{1}{c|}{}                                                                                     & Sentence count         & Count of unique words & Count of hashtags           \\
\multicolumn{1}{c|}{}                                                                                     & Count of mentions      & Count of stopwords    &                             \\ \midrule
\multicolumn{1}{c|}{\multirow{2}{*}{\begin{tabular}[c]{@{}c@{}}Similarity between \\ Text Pair\end{tabular}}} & SimCse                 & MPNet                 & The Fuzz                    \\
\multicolumn{1}{c|}{}                                                                                     & TF-IDF                 & Rouge                 &                             \\ \bottomrule
\end{tabular}
\caption{Selected features in our Feature Selector are 11 prevalent features in NLP and 5 common metrics for semantic textual similarities.}
\label{tab:Feature Selection}
\end{table}

Our initial focus involves computing common features in NLP for our datasets. Building on the methodology outlined in \cite{Feature_Engineering}, we calculate 11 features from each given text. These features include word length, the count of capital words, stopwords, quotes, and more. As our text sources encompass claim, evidence, question, claim answer to the question, and evidence answer to the question, we utilize these 11 metrics to comprehensively evaluate each text source. This results in a total of 55 features in the initial phase, significantly enabling our model to capture basic information distilled from the text.

Furthermore, we delve into the calculation of sentence similarity. Capitalizing on the flexibility afforded by various measurements of semantic and text similarities, we employ SimCSE, MPNet, The Fuzz, TF-IDF, and ROUGE to assess the similarity of given text pairs. Specifically, we evaluate the similarity between claim-evidence pairs and their corresponding question-answer pairs, resulting in a total of 10 features.


In summary, our approach involves the computation of essential text information and the evaluation of sentence similarity, resulting in a total of 65 features. Subsequently, we normalize these features and transform them into embeddings, denoted as $E_{FE}$.


\paragraph{Classifier}

The classifier is implemented as a simple two-layer Multilayer Perceptron (MLP). We directly concatenate the embeddings from the Pre-trained LLMs $E_{PLM}$ and the Feature Extractor $E_{FE}$, using the concatenated result as input for the classifier. Finally, the classifier produces the probability distribution over each label $\hat{y}_i$, where $W^{Z1} \in \mathbb{R}^{(d_{PLM} + d_{FE}) \times 128}$ and $W^{Z2} \in \mathbb{R}^{128 \times 3}$.

\begin{equation}
    \hat{y_i} = softmax((\sigma((E_{PLM} + E_{FE}) W^{Z1})) W^{Z2}),
\end{equation}

The loss function is cross-entropy and is defined as follows:

\begin{equation}
    \mathbb{L} = -\sum_{i=1}^{|C|} y_i log(\hat{y}_i)
\end{equation}

For further details on FakeNet, please refer to \cite{du2023team}.

\subsubsection{Fine-tuning Large Language Models (LLMs)}

Conversely, we explore an alternative approach by fine-tuning the entire pre-trained LLMs on the text classification task. In this configuration, the input is a sequence formed by concatenating the claim, evidence, question, claim answer, and evidence answer. The output of the model is the probability distribution over the three labels: Support, Neutral, or Refute.



\begin{equation}
\begin{matrix}
I = C_i + E_i + Q_i + CA_i + EA_i \\
\hat{y}_i = LLM(I)
\end{matrix}
\end{equation}

\subsubsection{Ensemble} \label{Ensemble}

Inspired by the approach outlined in \cite{du2023team}, we incorporate ensemble learning to amalgamate informative knowledge from various models, aiming to enhance the overall predictive performance. Four distinct ensemble methods have been designed:

\begin{enumerate}
    \item \textbf{Weighted sum with labels}: 
    \begin{equation}
    \begin{matrix}
        P_{s}' = W_{s1} \times P_{s1} + (1-W_{s}) \times P_{s2} \\
        P_{n}' = W_{n1} \times P_{n1} + (1-W_{n}) \times P_{n2} \\
        P_{r}' = W_{r1} \times P_{r1} + (1-W_{r}) \times P_{r2}
    \end{matrix}
    \end{equation}
    \item \textbf{Power weighted sum with labels}:
    \begin{equation}
    \begin{matrix}
        P_{s}' = W_{s1} \times P_{s1}^{E_{s}} + (1-W_{s}) \times P_{s2}^{\frac{1}{E_{s}}} \\
        P_{n}' = W_{n1} \times P_{n1}^{E_{n}} + (1-W_{n}) \times P_{n2}^{\frac{1}{E_{n}}} \\
        P_{r}' = W_{r1} \times P_{r1}^{E_{r}} + (1-W_{r}) \times P_{r2}^{\frac{1}{E_{r}}}
    \end{matrix}
    \end{equation}
    \item \textbf{Power weighted sum with two models}:
    \begin{equation}
    \begin{matrix}
    P' = W_{1} \times P_{1}^{E_{1}} + W_{2} \times P_{2}^{E_{2}}
    \end{matrix}
    \end{equation}
    \item \textbf{Power weighted sum with three models}:
    \begin{equation}
    \begin{matrix}
    P' = W_{1} \times P_{1}^{E_{1}} + W_{2} \times P_{2}^{E_{2}} + W_{3} \times P_{3}^{E_{3}}
    \end{matrix}
    \end{equation}
\end{enumerate}

In methods 1 and 2, the terms $P_{s}'$, $P_{n}'$, and $P_{r}'$ denote the new probabilities for Support, Neutral, and Refute, respectively. Meanwhile, $P_{si}$, $P_{ni}$, and $P_{ri}$ represent the probabilities for Support, Neutral, and Refute obtained from model $i$. In methods 3 and 4, the notation $P'$ signifies the new probabilities for all labels, while $P_{i}$ denotes the probabilities for all labels derived from model $i$.

The experimental results will be shown in Section \ref{Text Classification}.

\section{Experiments}

\subsection{Experiments Environment}

All experiments were executed on a machine equipped with 32 AMD EPYC 7302 16-Core CPUs, 2 NVIDIA RTX A5000 GPUs, and 252GB of RAM. Python serves as our primary programming language. The source code is available at https://github.com/AndyChiangSH/Pre-CoFactv3.

\subsection{Dataset}


We utilize the dataset FACTIFY5WQA \cite{surya2024factify}\cite{rani-etal-2023-factify} provided by the AAAI-24 Workshop Factify 3.0. This dataset is designed for fact verification, with the task of determining the veracity of a claim based on given evidence. In our specific task, we augment this dataset by introducing questions that entail the answers derived from the claim and evidence, thus framing the fact verification as an entailment problem. The dataset consists of 10,500 samples for training, 2,250 samples for validation, and 2,250 samples for testing, totaling 15,000 samples. Each sample in the training and validation sets includes fields for claim, evidence, question, claim answer, evidence answer, and label. For the testing set, the fields include claim, evidence, and question. 


\begin{itemize}
    \item \textbf{Claim}: the statement to be verified.
    \item \textbf{Evidence}: the facts to verify the claim.
    \item \textbf{Question}: the questions generated from the claim by the 5W framework (who, what, when, where, and why).
    \item \textbf{Claim answer}: the answers derived from the claim.
    \item \textbf{Evidence answer}: the answers derived from the evidence.
    \item \textbf{Label}: the veracity of the claim based on the given evidence, which is one of three categories: Support, Neutral, or Refute.
\end{itemize}

\subsection{Evaluation Metric}

The official competition metric for Factify 3.0 involves computing the average BLEU score for answers to questions derived from both the claim and evidence. A prediction is deemed correct only if this score exceeds a predefined threshold and the label is accurate. The final accuracy is then calculated as the percentage of correct predictions.

Following this, in the Question Answering task, we employ the BLEU score with 4-gram precision \cite{papineni2002bleu}. In the case of the Text Classification task, accuracy serves as the evaluation metric. Detailed experimental results are provided in the ensuing sections.
\subsection{Question Answering} \label{Question Answering}

\subsubsection{Fine-tuning Large Language Models (LLMs)}

\paragraph{Compare LLMs}

In this experiment, we aim to compare the results across different LLMs, namely:

\begin{enumerate}
    \item The roberta-large \cite{liu2019roberta} model fine-tuned by the SQuAD 2.0 \cite{2016arXiv160605250R} \footnote{https://huggingface.co/deepset/roberta-large-squad2} 
    \item The deberta-v3-large \cite{he2021debertav3} model fine-tuned by the SQuAD 2.0 \cite{2016arXiv160605250R} \footnote{https://huggingface.co/deepset/deberta-v3-large-squad2}
    \item The roberta-large \cite{liu2019roberta} model fine-tuned by the FACTIFY5WQA \cite{surya2024factify}.
    \item The deberta-v3-large \cite{he2021debertav3} model fine-tuned by the FACTIFY5WQA \cite{surya2024factify}.
    \item The roberta-large \cite{liu2019roberta} model fine-tuned by the SQuAD 2.0 \cite{2016arXiv160605250R} and the FACTIFY5WQA \cite{surya2024factify}.
    \item The deberta-v3-large \cite{he2021debertav3} model fine-tuned by the SQuAD 2.0 \cite{2016arXiv160605250R} and the FACTIFY5WQA \cite{surya2024factify}.
\end{enumerate}


Fine-tuning is conducted by the Hugging Face Trainer API on the Question Answering task\footnote{https://huggingface.co/docs/transformers/tasks/question\_answering}. We employ BLEU scores for both claim answer and evidence answer, taking the average of the two as the metric. The results are presented in Table \ref{tab:Compare LLMs}. Surprisingly, deberta-v3-large \cite{he2021debertav3} fine-tuned solely with the FACTIFY5WQA \cite{surya2024factify} dataset achieves the best performance. This outcome underscores the significance of fine-tuning on a dataset that matches the task's characteristics. Furthermore, it is worth noting that since a majority of the answers are short in length, they may yield lower BLEU scores with 4-gram precision. Finally, we use LLM 4 for testing.

\begin{table}[]
\centering
\begin{tabular}{c|ccc}
\toprule
\textbf{LLMs} & \textbf{Claim Answer (BLEU)} & \textbf{Evidence Answer (BLEU)} & \textbf{Average (BLEU)} \\ 
\midrule
1          & 0.3543          & 0.3006          & 0.3275          \\ 
2          & 0.3586          & 0.3178          & 0.3382          \\ 
3          & 0.5230          & 0.3361          & 0.4296          \\ 
\textbf{4} & 0.5248          & \textbf{0.3963} & \textbf{0.4605} \\ 
5          & \textbf{0.5323} & 0.3518          & 0.4421          \\ 
6          & 0.5268          & 0.3873          & 0.4571          \\ 
\bottomrule
\end{tabular}
\caption{Experiment result of Compare LLMs on the Question Answering task.}
\label{tab:Compare LLMs}
\end{table}

\subsection{Text Classification} \label{Text Classification}

In this section, we will present experimental results for the text classification task.

\subsubsection{FakeNet}

During the training of FakeNet, specific configurations were employed. The batch size was set to 24, the learning rate to 0.00005, and the number of epochs depended on the specific LLMs used. The input dimension for text and answer was 1024, the hidden dimension of FakeNet was set to 256, and the number of co-attention heads was specified as 2. Additionally, the input dimensions of the Pre-trained LLM $d_{PLM}$ and the Feature Extractor $d_{FE}$ were set to 256 and 32, respectively.

\paragraph{Compare Pre-trained LLMs}

To compare various Pre-trained LLMs, we employ seven LLMs available on Hugging Face. The results are displayed in Table \ref{tab:Compare Pre-trained LLMs}.

\begin{table}[]
\centering
\begin{tabular}{c|c|c}
\toprule
\textbf{Pre-trained LLMs}           & \textbf{Epoch} & \textbf{Accuracy} \\
\midrule
bert-large-uncased \cite{devlin2018bert}                   & 20             & 0.7040            \\
gpt2 \cite{radford2019language}                            & 30             & 0.6813            \\
t5-large \cite{raffel2020exploring}                        & 10             & 0.6991            \\
microsoft/deberta-large \cite{he2020deberta}               & 20             & 0.7498            \\
microsoft/deberta-xlarge  \cite{he2020deberta}             & 20             & 0.7440            \\
microsoft/deberta-v3-base \cite{he2021debertav3}           & 15             & 0.7364            \\
\textbf{microsoft/deberta-v3-large} \cite{he2021debertav3} & 15             & \textbf{0.7542}   \\
\bottomrule
\end{tabular}
\caption{Experiment results of different Pre-trained LLMs in FakeNet.}
\label{tab:Compare Pre-trained LLMs}
\end{table}


Two noteworthy findings have emerged from our experiments. Firstly, the model size does not exhibit a strictly positive correlation with performance. For instance, deberta-v3-large outperforms deberta-v3-base, but deberta-xlarge does not surpass deberta-large. Secondly, deberta-v3-large achieves the highest accuracy, prompting its selection for the next experiment.

\paragraph{Compare Features}

In the ensuing experiment, our objective is to assess whether the inclusion of features can enhance performance. We select 65 features as input and apply various normalization techniques. The detailed results are presented in Table \ref{tab:Compare Features}.

\begin{table}[]
\centering
\begin{tabular}{c|c}
\toprule
\textbf{Features}                             & \textbf{Accuracy} \\
\midrule
no features                                   & 0.7542             \\
features not normalized                       & 0.5880             \\
features normalized between 0 and 1           & 0.7502             \\
\textbf{features normalized between -1 and 1} & \textbf{0.7556}    \\
\bottomrule
\end{tabular}
\caption{Comparing features and various normalizations.}
\label{tab:Compare Features}
\end{table}

Our findings reveal that the inclusion of features normalized between -1 and 1 results in the best performance, albeit with a slight improvement of 0.19\%.

\subsubsection{Fine-tuning Large Language Models (LLMs)}

\paragraph{Compare Input and Length}


In the fine-tuning process, we employ the microsoft/deberta-v3-large \cite{he2021debertav3} \footnote{https://huggingface.co/microsoft/deberta-v3-large} as the Pre-trained LLMs. The training parameters are configured with an epoch of 8, a batch size of 4, and a learning rate of 0.00002. Fine-tuning is executed using the Hugging Face Trainer API on the Text Classification task \footnote{https://huggingface.co/docs/transformers/tasks/sequence\_classification}, and various alterations in the input and length are made to facilitate performance comparison. The result is shown in Table \ref{tab:Compare input text and length}.

\begin{table}[]
\centering
\resizebox{\columnwidth}{!}{%
\begin{tabular}{c|ccccc|c}
\toprule
\textbf{Input} & \textbf{Claim} & \textbf{Evidence} & \textbf{Question} & \textbf{Evidence Answer} & \textbf{Claim Answer} & \textbf{Accuracy} \\
& \textbf{Length} & \textbf{Length} & \textbf{Length} & \textbf{Length} & \textbf{Length} \\
\midrule
text                     & 100  & 1000  & -  & -  & -   & 0.8044          \\
text                     & 400  & 4000  & -  & -  & -   & 0.8396          \\
text                     & 800  & 8000  & -  & -  & -   & 0.8462          \\
text                     & 1600 & 10000 & -  & -  & -   & \textbf{0.8502} \\
question + answer        & -    & -     & 50 & 50 & 100 & 0.6311          \\
text + question + answer & 100  & 1000  & 50 & 50 & 100 & 0.7849          \\
\bottomrule
\end{tabular}%
}
\caption{The performance comparison between different alterations in the input and length.}
\label{tab:Compare input text and length}
\end{table}


Surprisingly, the performance with only text as input surpasses that of using text, question, and answer as input. This prompts a question: Is the inclusion of question and answer truly beneficial for text classification? Additionally, we observe that longer text lengths correspond to improved performance. The model with a claim length of 1600 and evidence length of 10000 achieves the highest accuracy, markedly surpassing FakeNet.

\subsubsection{Ensemble}

The experimental results for the four ensemble methods mentioned in Section \ref{Ensemble} are provided in Table \ref{tab:Ensemble}.

\begin{table}[]
\centering
\resizebox{\columnwidth}{!}{%
\begin{tabular}{c|ccc|c}
\toprule
\textbf{Ensemble Methods}                     & \textbf{Model 1} & \textbf{Model 2} & \textbf{Model 3} & \textbf{Accuracy} \\
\midrule
\textbf{Weighted sum with labels}           & Fine-tuned LLM 1 & Fine-tuned LLM 2 & - & 0.8564 \\
\textbf{Power weighted sum with labels}     & Fine-tuned LLM 1 & Fine-tuned LLM 2 & - & 0.8587 \\
\textbf{Power weighted sum with two models} & Fine-tuned LLM 1 & Fine-tuned LLM 2 & - & 0.8609 \\
\textbf{Power weighted sum with three models} & Fine-tuned LLM 1 & Fine-tuned LLM 2 & FakeNet          & \textbf{0.8644}  \\
\bottomrule
\end{tabular}%
}
\caption{The experimental results for the four ensemble methods, where Fine-tuned LLM 1 is the fine-tuned LLM with a claim length of 800 and evidence length of 8000, Fine-tuned LLM 2 is the fine-tuned LLM with a claim length of 1600 and evidence length of 10000, and FakeNet is the deberta-v3-large training with features normalized between -1 and 1.}
\label{tab:Ensemble}
\end{table}



Our examination indicates that the ensemble method, utilizing a power weighted sum with three models (two Fine-tuned LLMs and one FakeNet), achieves the highest accuracy. Hence, we have selected this ensemble model as our best result.

\begin{figure}
  \centering
  \includegraphics[width=\linewidth]{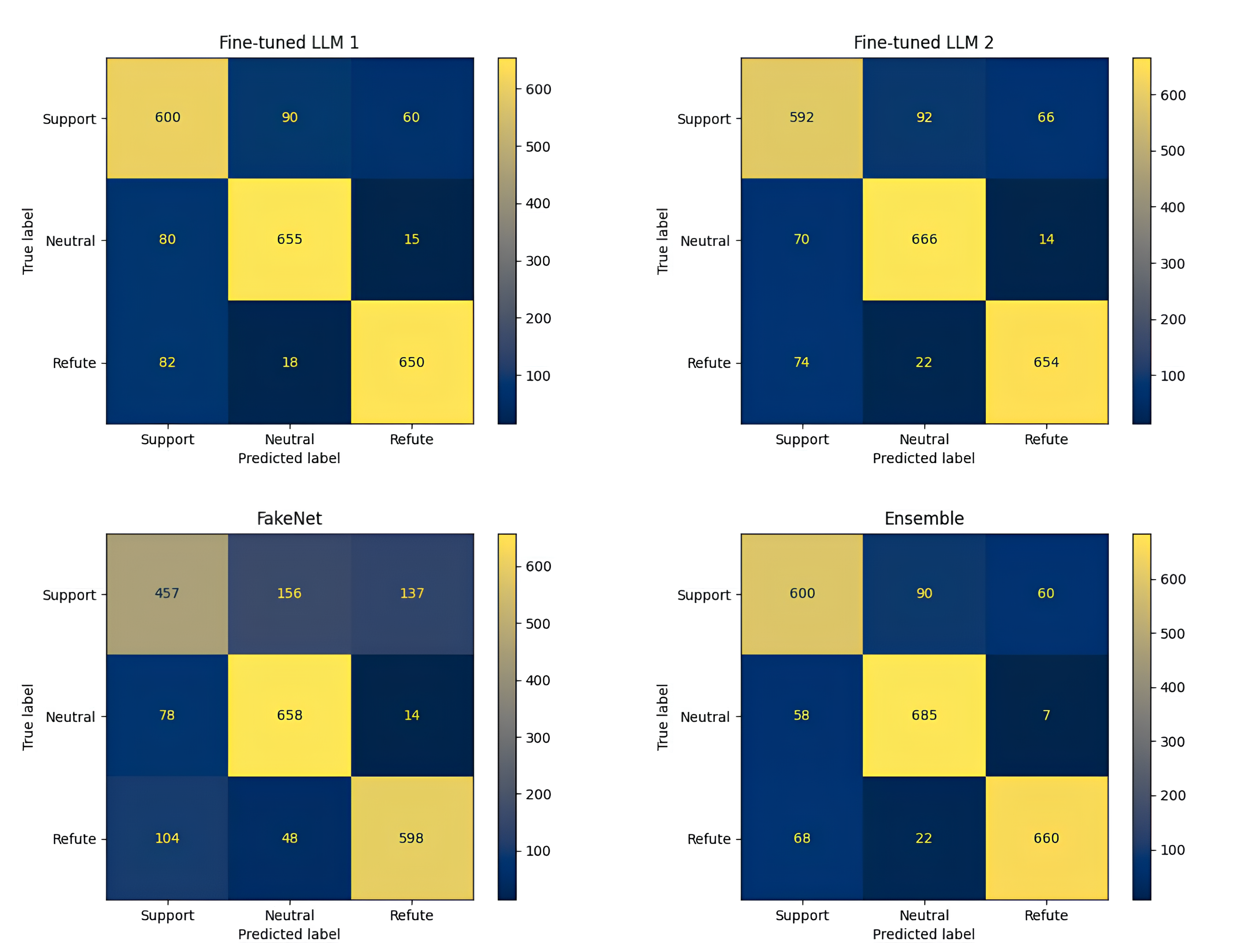}
  \caption{The confusion matrix for these three models and the ensemble model}
  \label{fig:confusion_matrix}
\end{figure}

Figure \ref{fig:confusion_matrix} displays the confusion matrix for these three models and the ensemble model. The ensemble approach demonstrates the ability to capitalize on the strengths of each individual model, leading to an overall enhancement in accuracy. However, it is noteworthy that all models exhibit difficulty in accurately identifying instances of "Support," thereby limiting the extent of performance improvement in this particular category.

\subsection{Baselines}

\subsubsection{In-Context Learning Baseline}
The In-Context Learning baseline uses the ChatGPT API to generate replies with the best prompt we could come up with, more details about prompt engineering in \ref{In-Context Learning Text Classification}. Around one hundred claim-evidence pairs were fed into the Large Language Model, and the respective labels were extracted from their replies.
The result in Table \ref{tab:baseline} seemed skewed towards support, but after further analysis, we found it rather difficult to differentiate between support and neutral, even as humans. We take the lower percentages of correct neutral labels to mean the prompt we were using favored support when in an ambiguous state. Overall, it does not outperform the human baseline, and falls way short of our Pre-CoFactv3 model.

\subsubsection{Human Baseline}

To assess the effectiveness of our model, we established a human baseline through a questionnaire created using the dataset. The objective was to gauge how well our model performs compared to human beings in identifying fake news. For simplicity, we selected 10 entities from the dataset and distributed them to 20 individuals, asking them to categorize the claim news. The results are presented in the Table \ref{tab:baseline}.

\begin{table}[]
\begin{tabular}{@{}c|cccc@{}}
\toprule
 & \textbf{Support} & \textbf{Neutral} & \textbf{Refute} & \textbf{Total}  \\ 
\midrule
In-Context Learning Baseline & 0.7500 & 0.2857 & 0.3333 & 0.4300 \\
Human Baseline & 0.5500  & 0.6000  & 0.1333 & 0.4400 \\
\midrule
\midrule
\textbf{Pre-CoFactv3} & \textbf{0.8000} & \textbf{0.9133}  & \textbf{0.8800} & \textbf{0.8644} \\
\bottomrule
\end{tabular}
\caption{Accuracy of Human Baseline, In-Context Learning Baseline, and Pre-CoFactv3 with different labels.}
\label{tab:baseline}
\end{table}

Upon analysis, the accuracy of human identification was found to be 44\%, significantly lower than our model's performance. Moreover, it became apparent that, concerning the Refute metric, a significant number of participants struggled to provide correct answers. This challenge likely stems from the ambiguous boundary between Neutral and Refute metric. This observation serves as a foundation for illustrating the efficacy of our model in surpassing human judgment in fake news identification.

\subsection{Testing Result}

\subsubsection{Final Submission}


In our final submissions, we present three versions of our work, and the corresponding testing accuracy is provided in Table \ref{tab:Final Submission}. It is noteworthy that the performance of the ensemble model did not meet our expectations, possibly indicating overfitting issues. Given that the fine-tuned LLM for both Question Answering and Text Classification demonstrates the highest testing accuracy, our observation leads us to assert that "Fine-tuning is all you need."

\begin{table}[]
\centering
\begin{tabular}{c|cc|c}
\toprule
\textbf{Submission} & \textbf{Question Answering} & \textbf{Text Classification} & \textbf{Testing Accuracy} \\
\midrule
1 & Fine-tuned LLM & FakeNet        & 0.6880          \\
2 & Fine-tuned LLM & Fine-tuned LLM & \textbf{0.6956} \\
3 & Fine-tuned LLM & Ensemble       & 0.6080          \\
\bottomrule
\end{tabular}
\caption{The testing accuracy of three submissions.}
\label{tab:Final Submission}
\end{table}

\subsubsection{Leaderboard}


Table \ref{tab:Leaderboard} displays the leaderboard for the Factify 3.0 Workshop. We are delighted to announce that our team, Trifecta, won first place in the workshop. Our performance surpassed the baseline by an impressive 103\%, maintaining a lead over the second competitor by 70\%.

\begin{table}[]
\centering
\begin{tabular}{c|c}
\toprule
\textbf{Team Name}       & \textbf{Testing Accuracy (\%)} \\
\midrule
Baseline                 & 0.3422 (0\%)                   \\
Jiankang Han             & 0.4547 (33\%)                  \\
SRL\_Fact\_QA            & 0.4551 (33\%)                  \\
\midrule
\midrule
\textbf{Trifecta (Ours)} & \textbf{0.6956 (103\%)}        \\
\bottomrule
\end{tabular}
\caption{The leaderboard for the Factify 3.0 Workshop.}
\label{tab:Leaderboard}
\end{table}

\section{Limitations \& Discussions}

Fine-tuning a model demands a wealth of annotated data for effective supervised training, and any shortfall in labeled data can detrimentally impact the training outcomes. Moreover, the current dataset is confined to English data, restricting its applicability across different languages. 

The real-world data introduces added complexities, with nuances in semantics and incomplete information posing challenges to the model's recognition capabilities, such as using a pre-built database of evidence to fact-check claims made on social media and other information-sharing platforms. While this brings into question whether previously collected evidence could confirm the news as it happens, it could very well be used to prune incoming data, lessening the load for further examination, whether by more complex models or human intervention.

In the fine-tuning process, parameters like model size, max length, and batch size are intricately tied to GPU memory constraints. Exploring options such as expanding GPU memory or adopting Parameter-Efficient Fine-Tuning (PEFT) raises intriguing questions for future research, particularly in relation to their potential to amplify recognition accuracy.




\section{Conclusion}





In this paper, we introduced \textbf{Pre-CoFactv3}, a framework composed of two integral parts: (1) Question Answering and (2) Text Classification. For the Question Answering task, we employed In-Context Learning and Fine-tuned Large Language Models (LLMs) to answer questions based on the claim and evidence. In the Text Classification task, our approach encompassed In-Context Learning, the FakeNet model, and Fine-tuned LLMs to predict label categories.

Our experiments demonstrated the efficacy of our diverse approaches, encompassing the comparison of various Pre-trained LLMs, the development of the FakeNet model, and the exploration of different ensemble methods. Additionally, we established two baseline models for In-Context Learning and Human performance.

The culmination of our efforts resulted in the success of our team, Trifecta, securing the first-place position in the AAAI-24 Factify 3.0 Workshop. Our performance surpassed the baseline by an impressive 103\%, maintaining a substantial lead over the second competitor by 70\%.

In summary, our accomplishment validates the effectiveness of our approach and the thoughtful integration of diverse techniques. As we continue to advance in the field of fact verification, our experiences offer valuable insights and lessons for the broader research community.


\bibliography{references}

\appendix

\section{In-Context Learning}

\subsection{Question Answering} \label{In-Context Learning Question Answering}

\subsubsection{Prompt Designs}

We experimented with In-Context Learning for Question Answering task using GPT-3.5 \cite{ChatGPT}. The following system prompt $P_s$ was engineered in a way we believed would unlock its full potential, combining methods of characterization, scope limitation and response formatting. \\ \newline
$P_s$ = "You are a master of question-answering. 
I will give you the context and question. 
You can ONLY answer the question based on the context. 
The answers not included in the context are not allowed!
Answer the questions as short as possible.
Now, please answer the questions below."

\subsubsection{Experimental Results}

Prompts were then given in the context-question-answer format $F = (C_l + C_t) + (Q_l + Q_t) + (A_l)$, where $l$ and $t$ denotes label and text respectively, leaving $A_t$ for GPT-3.5 to generate. The results are presented in Table \ref{tab:ICL-QA-result}.


\begin{longtable}{l|p{2.5cm}p{2.5cm}|p{2.5cm}p{2.5cm}}
\toprule
\textbf{Category} & \textbf{Claim} & & \textbf{Evidence} & \\ \midrule
\textbf{Method} & \multicolumn{1}{l}{\textbf{GPT-3.5}} & \textbf{Ground Truth} & \multicolumn{1}{l}{\textbf{GPT-3.5}} & \textbf{Ground Truth} \\ \midrule
Question 1 & \multicolumn{1}{p{3cm}}{Micah Richards.}        & Micah Richards              & \multicolumn{1}{p{3cm}}{The context does not specify anyone spending an entire season at Aston Villa without playing a single game.}        & Richards             \\ \midrule
Question 2 & \multicolumn{1}{p{3cm}}{Zero.}        & a single game             & \multicolumn{1}{p{3cm}}{The context does not provide information on the number of games Micah Richards played at Aston Villa.}        & 0             \\ \midrule
Question 3 & \multicolumn{1}{p{3cm}}{Aston Villa.}        & : at Aston Vila             & \multicolumn{1}{p{3cm}}{Micah Richards spent an entire season at Aston Villa without playing a game.}        & Aston Villa             \\ \midrule
Question 4 & \multicolumn{1}{p{3cm}}{The context does not specify the exact duration of Micah Richards' time at Aston Villa without playing.}        & an entire season             & \multicolumn{1}{p{3cm}}{The context does not specify the duration of Micah Richards' time at Aston Villa.}        & remained at the club             \\ 
\bottomrule
\caption{Table of comparison answers between GPT-3.5 and ground truth in question answering task.}
\label{tab:ICL-QA-result}
\end{longtable}

We found that GPT-3.5 was not completely capable of responding verbatim with answers taken from the context. Since our method of evaluation was of BLEU score \cite{papineni2002bleu} with answers that were taken directly from the context, In-Context Learning was ultimately disregarded as a viable method for solving Question Answering task. 

\subsection{Text Classification} \label{In-Context Learning Text Classification}

\subsubsection{Prompt Designs}
We experimented with a large amount of prompt templates to find one that best fit the task of text classification using GPT-3.5 \cite{ChatGPT}. Ten of the most noteworthy designs are as Table \ref{tab:prompt-template}: 

\begin{longtable}{ll|p{9cm}}
\toprule
\textbf{ID} & \textbf{Descriptor} & \textbf{Prompt} \\ \midrule
1  & Zero-Shot           & System Prompt: given claim text and evidence text, determine whether the evidence refutes, supports or is neutral against the claim.\\ \midrule
2  & Few-Shot \cite{brown2020language}           & System Prompt: given claim text and evidence text, determine the probabilities that the evidence supports, refutes and is neutral against the claim. At the end, return a label with the highest probability. Here are examples of each as the highest probability:\newline1.\newline "claim": "The Washington Metro has a frequency of 4 minutes on the Red Line and 9 minutes on other lines , during rush hour .",\newline"evidence": "Trains run more frequently during rush hours on all lines , with scheduled peak hour headways of 3 minutes on the Red Line and 6 minutes on all other lines .",\newline "label": "Refute"\newline2.\newline "claim": "Scientists have developed a more accurate way to determine dogs' ages, rather than multiplying human years by seven. https://t.co/u4E7BJuQ4U",\newline"evidence": "By Francesca Giuliani-Hoffman, CNNUpdated 2:57 PM ET, Sat July 4, 2020 (CNN)How do you compare a dog's age to that of a person? A popular method says you should multiply the dog's age by 7 to compute how old Fido is in \"human years. \"",\newline "label": "Support"\newline3.\newline "claim": "More than 400,000 copies of Good Girl Gone Bad were shipped to different parts of the world .",\newline"evidence": "It was certified quintuple platinum by Music Canada , denoting shipments of more than 500,000 copies .",\newline "label": "Neutral"\newline Please answer to the best of your abilities. \\ \midrule
3  & Probability         &System Prompt: given claim text and evidence text, determine the probabilities that the evidence supports, refutes and is neutral against the claim.  \\ \midrule
4  & Summarized             &System Prompt: given claim text and evidence text, summarize the evidence first, then determine the percentage-wise probabilities that the evidence supports, refutes, and is neutral against the claim. Finish with either \{Support\}, \{Refute\} or \{Neutral\}. \\ \midrule
5  & Defined &System Prompt: given claim text and evidence text, determine the probabilities that the evidence supports, refutes and is neutral against the claim. Support means both are related and is likely true. Refute means both are related but is likely false. Neutral means both are unrelated.        \\ \midrule
6  & Characterized       &System Prompt: Pretend you are a human fact checker tasked with the following: given claim text and evidence text, determine the probabilities that the evidence supports, refutes and is neutral against the claim. \\ \midrule
7  & Chain of Thought \cite{NEURIPS2022_9d560961}    &System Prompt: given claim text and evidence text, summarize the evidence first, then determine the percentage-wise probabilities that the evidence supports, refutes and is neutral against the claim. Remember to walk us through the thought process. Finish with either \{Support\}, \{Refute\} or \{Neutral\}. \\ \midrule
8  & Socratic \cite{chang2023prompting}            &System Prompt: given a claim text and evidence text, find five reasons from evidence that supports claim and five reasons from evidence that refutes claim. Give each reason a likelihood probability and how important it is to the validity of the claim. Multiply the two values with each other, turn the refute reasons negative, and add all six reasons together. Return {Support} if it is higher than 0.1, or {Refute} if it is lower than -0.1, or {Neutral} if it is in between. \\ \midrule
9  & With QA             &System Prompt: given claim, evidence, question, answer to question by claim, answer to question by evidence, from 0 to 100, give probabilities that the evidence supports, refutes or is neutral against the claim. Finish the response with the most likely choice overall in brackets {}. \\ \midrule
10 & QA Only             &System Prompt: given question, answer to question by claim, answer to question by evidence, from 0 to 100, give probabilities that the evidence supports, refutes or is neutral against the claim. Finish the response with the most likely choice overall in brackets {}. \\ 
\bottomrule
\caption{Table of prompt template candidates to represent In-Context Learning for text classification tasks.}
\label{tab:prompt-template}
\end{longtable}

\subsubsection{Experimental Results}
Experiments were carried out by checking which prompt template could predict the correct label and whether it could predict the correct label in instances where others could not. Table \ref{tab:prompt-template-test} is a brief overview of how each prompt template performed when given identical tasks. "O" indicates a correct label was predicted, "X" indicates a correct label was not predicted, and "-" indicates there was no need for further testing since it does not outperform other prompt templates.

\begin{longtable}{ll|ccc}
\toprule
\textbf{ID} & \textbf{Descriptor} & \textbf{Test 1} & \textbf{Test 2} & \textbf{Test 3} \\ \midrule
1  & Zero-Shot           & X      & -      & -      \\ \midrule
2  & Few-Shot            & -      & -      & X      \\ \midrule
3  & Probability         & O      & O      & X      \\ \midrule
4  & Summarized          & X      & -      & X      \\ \midrule
5  & Defined             & -      & -      & X      \\ \midrule
6  & Characterized       & -      & -      & X      \\ \midrule
7  & Chain of Thought    & X      & -      & -      \\ \midrule
8  & Socratic            & -      & -      & X      \\ \midrule
9  & With QA             & X      & -      & -      \\ \midrule
10 & QA Only             & X      & -      & -      \\
\bottomrule
\caption{Table of prompt templates and their performance on test instances of text classification.}
\label{tab:prompt-template-test}
\end{longtable}

Although some prompt templates may perform at the same level, when taking into consideration the cost and convenience of execution, "Probability" seemed to have the highest competency of them all, so is ultimately chosen as the final baseline for comparison with our own models. Similar prompt templates were also tested on Llama 2 \cite{touvron2023llama}, but we did not find an increase in accuracy when compared to GPT-3.5.
The In-Context Learning baseline was then created by feeding instances of text classification task through GPT-3.5 using the "Probability" prompt template. After comparing the output with the correct labels, its accuracy was recorded at 43 percent, close to what others have achieved on classification tasks with In-Context Learning \cite{min2022rethinking}.

\end{document}